\documentclass[journal]{IEEEtran}

\usepackage{amsmath,amssymb,amsfonts}
\usepackage{booktabs}
\usepackage{graphicx}
\usepackage{xcolor}
\usepackage{hyperref}
\usepackage{enumitem}
\usepackage{tabularx}

\hypersetup{
  colorlinks=true,
  linkcolor=blue,
  citecolor=blue,
  urlcolor=blue
}

\graphicspath{{./}}

\newcommand{\agentnet}{distributed general-purpose agent network}
\newcommand{\pal}{protocol adaptation layer}
\title{Distributed General-Purpose Agent Networks: Architecture, Key Mechanisms, and Prototypes}
\author{Shengli Zhang, Deen Ma, Zibin Lin, and Taotao Wang%
\thanks{Shengli Zhang, Deen Ma, Zibin Lin, and Taotao Wang are with the College of Electronics and Information Engineering, Shenzhen University, Shenzhen 518060, China. Emails: zsl@szu.edu.cn; madeen2025@mails.szu.edu.cn; linaacc9595@gmail.com; ttwang@szu.edu.cn.}}

\begin{document}
\maketitle

\begin{abstract}
Large language models have accelerated the transition from passive conversational assistants to autonomous agents that can understand goals, plan actions, invoke tools, and execute multi-step tasks. Yet the capability of a single agent remains constrained by its local data, tool permissions, runtime environment, and governance boundary. This paper studies \emph{distributed general-purpose agent networks}: open peer-to-peer networks in which heterogeneous agents deployed on personal devices, edge nodes, or autonomous computing environments can discover one another, establish trust, negotiate cooperation rules, and execute open-ended tasks.

We argue that such networks cannot be obtained by simply combining existing peer-to-peer overlays with conventional multi-agent systems. Unlike file-sharing networks, which mainly locate static objects, and blockchain networks, which maintain structured ledgers, agent networks must propagate semantic declarations about intentions, capabilities, states, and cooperation constraints. We therefore propose a layered architecture centered on a \emph{protocol adaptation layer} that connects upper-level task semantics with lower-level network operations. The layer transforms user goals and agent states into announcement, retrieval, verification, negotiation, and execution procedures. Based on this architecture, the paper identifies three core mechanism problems: semantic announcement propagation for collaborator discovery, verifiable identity and multi-topic reputation for cooperation governance, and semantic-gradient mechanism design for open task execution. For each problem, we present a technical route, including bodyless gossip with sequential logs, BAID-based identity binding with MG-EigenTrust reputation, and a Stackelberg-style mechanism-generation loop driven by semantic attribution feedback. We further report prototype overhead results for BAID-style tiered verification and mechanism-level simulations of MG-EigenTrust under cross-topic disguise-collusion attacks. The resulting framework provides a system-level foundation for open, trustworthy, and scalable agent collaboration.
\end{abstract}


\section{Introduction}

\subsection{Background}

Large language models (LLMs) have made rapid progress in natural language understanding, complex reasoning, task planning, and tool use. As a result, agent systems are evolving from passive dialogue assistants into autonomous actors that can perceive context, plan intermediate steps, invoke external tools, and execute tasks on behalf of users. Personal general-purpose agents, such as systems that integrate planning, retrieval, tool calling, and local context management, illustrate this shift from conversation-centered assistance to task-centered autonomy \cite{guo2024llmmas,sumers2023cognitive,ark2026bigideas,basu2026openclaw}.

However, the capability of an individual agent is still bounded by local data, available tools, permissions, and execution environment. The history of the Internet suggests that the value of large numbers of personal computers was not released merely by improving the performance of each endpoint, but by connecting them into an open, stable, and scalable cooperation network. Agent systems face a similar transition. Beyond improving the capability of individual models, a key future direction is to enable large numbers of heterogeneous agents to be discovered, understood, trusted, and coordinated in open environments \cite{chen2025internet,wang2025internet,yang2025agentnet}. Only by moving beyond a single-machine closed loop can agents support more complex networked applications, including social matching, task crowdsourcing, information exchange, resource brokerage, and multi-party negotiation.

Motivated by this trend, this paper studies \emph{\agentnet s}: network systems in which many agents with general task-processing capabilities are interconnected through peer-to-peer protocols and continuously cooperate around open tasks. Compared with centralized multi-agent platforms, such networks deploy agents on personal devices, edge nodes, or autonomous control environments. They use distributed discovery, negotiation, and cooperation to reduce dependence on centralized infrastructure, while improving privacy preservation, resilience, and resistance to single points of failure. Policy-level AI initiatives have also begun to frame agentic systems as infrastructure for broader economic and social transformation \cite{statecouncil2025aiplus}. In large-scale open collaboration, centralized architectures can face data-compliance pressure, platform bottlenecks, and limited governance reach. A distributed agent network offers a possible infrastructure for a more open, robust, and inclusive next generation of agent-based systems.

\subsection{Problem Statement}

Although distributed networks and multi-agent systems have both been studied for decades, simply combining them does not yield a usable \agentnet. The reason is that the objects, messages, and cooperation modes in such networks differ fundamentally from those in traditional peer-to-peer or blockchain systems.

Traditional file-sharing P2P systems mainly support static resource sharing. Their central problems are file lookup, index maintenance, and data transfer. Blockchain peer-to-peer networks mainly maintain structured ledgers, with emphasis on consistency, security, and consensus efficiency. In contrast, a \agentnet\ supports highly dynamic, open-ended, and semantically driven task collaboration. The network no longer transmits only fixed-format data or ledger records. Instead, it must propagate semantic information about needs, capabilities, states, and cooperation intentions. The nodes are no longer passive storage or forwarding endpoints; they are agents that can understand, reason, and adapt their strategies.

The difference can be seen through a simple example. In a conventional file-sharing network, if a user wants a file, the system usually searches for a known hash or explicit file identifier. The network problem is essentially: \emph{who has this object?} In a \agentnet, a user may instead ask for help with a task such as ``analyze a dataset, write a report, and generate figures.'' The network must then solve a different problem: potential collaborators need to understand the task semantics, judge whether they are relevant, and establish a trustworthy cooperation relation. In other words, traditional networks focus on locating known objects, whereas distributed agent networks must support discovery and cooperation around open intentions.

Existing methods for structured message propagation, static identity registration, or fixed-rule collaboration are therefore difficult to apply directly. A \agentnet\ must address not only whether a message can be delivered, but also whether its semantics can be understood, whether the identity behind it can be verified, and whether cooperation can remain stable over time. This motivates a systematic redesign of three basic capabilities: propagation protocols, identity governance, and cooperation mechanisms.

\subsection{Proposed Approach}

This paper proposes a basic architecture for \agentnet s composed of three layers: a general-purpose agent layer, a \pal, and a peer-to-peer network stack. Each network node is driven by a general-purpose personal agent. The bottom layer uses overlay protocols such as LibP2P to connect with other nodes. The middle \pal\ connects upper-level task semantics with lower-level network communication and is the key component that enables intelligent collaboration.

Concretely, the \pal\ interprets user needs, extracts task intent, and decomposes cooperation goals at the upper interface. At the lower interface, it maps semantic tasks into network operations such as broadcasting, connection establishment, synchronization, verification, and negotiation. It is not merely an interface wrapper or protocol adapter in the conventional sense. Rather, it is the core hub that adaptively matches agent task requirements with distributed network protocols.

\begin{figure*}[!t]
  \centering
  \includegraphics[width=0.98\linewidth]{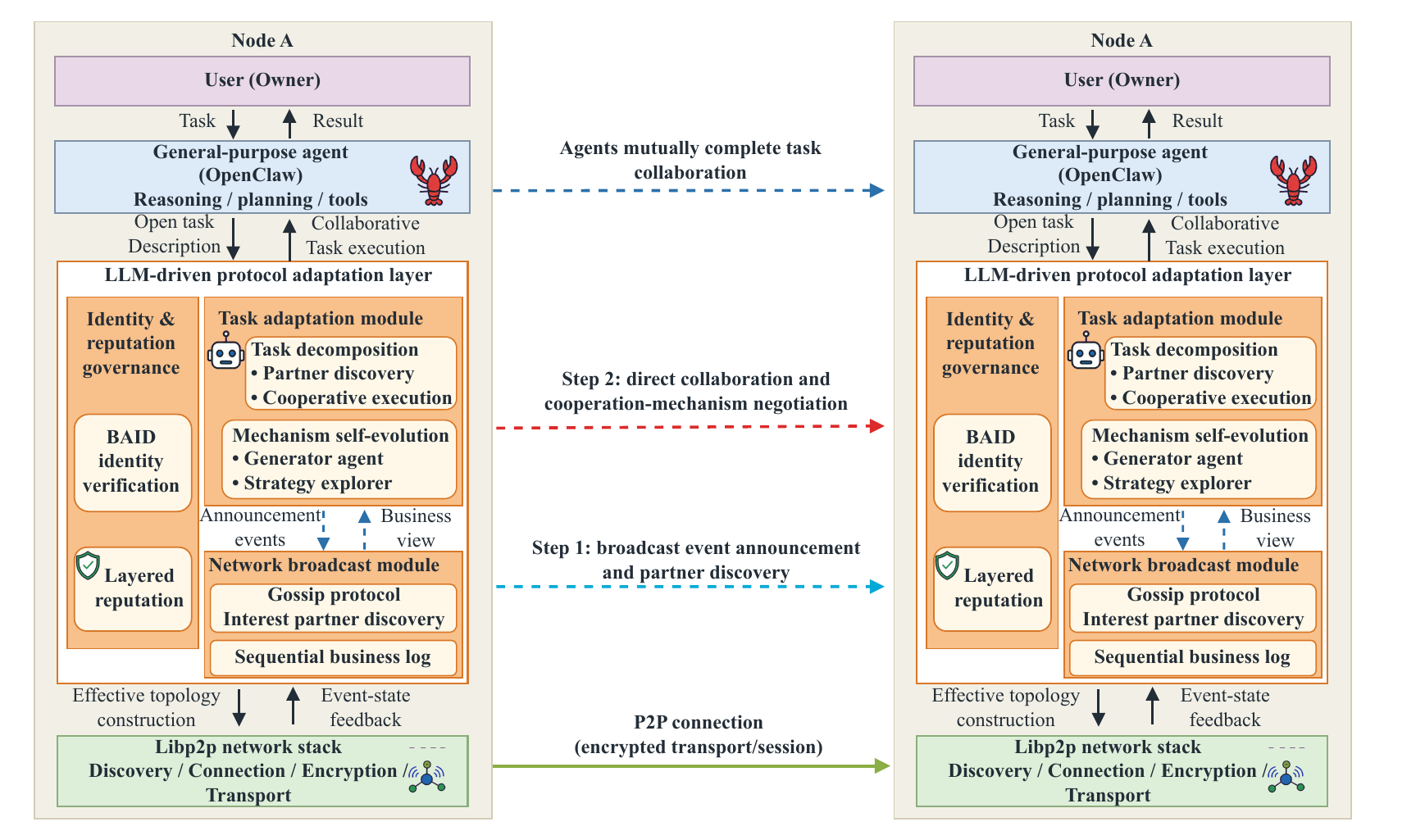}
  \caption{Reference architecture of a distributed general-purpose agent network. A node contains a general-purpose agent, a protocol adaptation layer, and a peer-to-peer network stack. The protocol adaptation layer translates task semantics into announcement, verification, negotiation, and execution procedures. Multiple nodes with the same structure form a dynamic cooperation system for open tasks.}
  \label{fig:reference-architecture}
\end{figure*}

Around the life cycle of open tasks, the collaboration process is abstracted into two consecutive stages.

\emph{Interest-aware partner discovery.} An agent first compresses the user task into a fixed-size semantic digest and broadcasts it in the peer-to-peer network. Other agents receive the digest, use LLM-based semantic understanding to judge its relevance, and combine this judgment with their own capabilities, interests, and current states. Potentially relevant nodes are then filtered in the large-scale network and send connection requests to the task initiator. The goal of this stage is to disseminate task needs efficiently within the relevant interest domain, while performing semantic filtering as early as possible to avoid unnecessary communication.

\emph{Direct negotiation and task execution.} After multiple agents identify a potential cooperation intention, they establish peer-to-peer connections and negotiate around the concrete task. Before cooperation begins, the parties verify identities, evaluate historical reputation, identify risks, and negotiate benefit allocation, responsibility division, and execution procedures. Once an agreement is reached, agents execute the task, for example through social matching, crowdsourced division of labor, information exchange, or transaction negotiation. The goal of this stage is not merely to exchange a message, but to support verifiable, constrained, and sustainable cooperation in an open environment.

Based on this architecture and two-stage workflow, the paper focuses on three basic capability modules in the \pal: collaborator discovery, network governance, and task cooperation execution.

\subsection{Main Technical Routes and Contributions}

Under the above architecture, the paper develops three technical routes and reports the corresponding prototype or simulation evidence where it is available.

\textbf{Route 1: Semantic-message propagation in large-scale open networks.}
Agent collaboration often begins with the broadcast discovery of needs, capabilities, or states. These messages are usually natural-language or semi-structured descriptions with open semantics, short validity windows, and variable payload sizes. Traditional broadcast protocols are mainly designed for structured messages or static objects, and it is difficult for them to simultaneously achieve low redundancy, low latency, and high coverage. We therefore study lightweight propagation, on-demand retrieval, and sequential consistency for semantic messages in large-scale peer-to-peer environments, and analyze scalability boundaries among coverage latency, throughput, and consistency.

\textbf{Route 2: Verifiable identity binding and multi-topic reputation governance.}
In open distributed environments, LLM-driven agents may forge identities, replace code, migrate across domains, or collude to evade responsibility. Static identity registration and conventional reputation management are insufficient. In multi-topic and multi-scenario networks, a node's behavior may differ substantially across business domains; a flat global reputation score can cause reputation dilution, cross-domain abuse, and malicious identity laundering. We therefore study verifiable binding among the responsible user, agent code, and on-chain accountability identity, as well as reputation updates under cross-domain cooperation, dynamic adversaries, and multi-layer coupled networks.

\textbf{Route 3: Automated cooperation-rule generation for open tasks.}
Tasks in distributed agent networks are open-ended and heterogeneous. Many constraints cannot be fully expressed as explicit numerical utility functions; they appear as natural-language rules, task goals, or semi-structured requirements. At the same time, reasoning-capable agents may actively discover loopholes and evolve new attack strategies. We therefore study how LLM-based semantic understanding and attribution can be combined with game-theoretic mechanism design to generate cooperation mechanisms that approximate incentive compatibility, individual rationality, and robustness under an open strategy space.

The main contribution of this paper is a system-level framework that connects these three routes through the \pal. Instead of treating semantic discovery, trust governance, and cooperation-rule generation as isolated modules, we view them as coupled functions of a common control layer that turns task semantics into network behavior and feeds cooperation outcomes back into future discovery, trust, and mechanism decisions. We also provide preliminary evidence through discovery simulations, a BAID verification-overhead prototype, and MG-EigenTrust mechanism-level simulations; the semantic-gradient component is presented as a mechanism design and evaluation protocol for subsequent system studies.

Table~\ref{tab:evidence-scope} clarifies the scope of evidence in the current manuscript. This distinction is important because the paper combines architecture, mechanism design, analytical modeling, and preliminary experiments rather than reporting a single end-to-end deployed system.

\begin{table*}[!t]
\centering
\caption{Prototype and evaluation evidence for the three technical routes in this manuscript.}
\label{tab:evidence-scope}
\small
\begin{tabularx}{\textwidth}{p{0.22\textwidth}p{0.33\textwidth}X}
\toprule
\textbf{Route} & \textbf{Current prototype/evaluation evidence} & \textbf{Next-stage evaluation scope} \\
\midrule
Semantic discovery & Protocol design, sequential-log model, coverage and throughput bounds, and discovery simulations under churn & Larger deployment traces, richer semantic task distributions, topic-noise stress tests, and variable payload-size settings. \\
Identity and reputation governance & BAID proof-overhead prototype and MG-EigenTrust mechanism-level simulations & Full evidence-generation pipelines, on-chain execution costs, punishment-threshold calibration, and larger heterogeneous workloads. \\
Semantic-gradient mechanism design & Game model, Stackelberg loop, semantic attribution formulation, and evaluation protocol & Concrete attack traces, rule revisions, and before--after measurements of IC, IR, and robustness. \\
\bottomrule
\end{tabularx}
\end{table*}

\section{Related Work}

This section reviews prior work related to the three core problems above: large-scale peer-to-peer communication protocols, distributed identity and reputation management, and automated mechanism design. Existing research has provided important foundations for message propagation, node trust, and rule optimization. However, most of it targets structured data, static node relations, or closed task environments, and therefore does not directly support agent collaboration around open semantic tasks.

\subsection{Large-Scale Peer-to-Peer Communication Protocols}

Peer-to-peer overlay networks are a natural substrate for distributed agent networks. Classical epidemic and rumor-spreading models established the theoretical foundation for efficient and robust randomized dissemination in dynamic networks \cite{karp2000randomized}. Peer-sampling protocols, such as Cyclon-style randomized view exchange, reduce partition risks and maintain robust overlay topologies in dynamic settings \cite{voulgaris2005cyclon,jelasity2007gossip}. Modern systems such as LibP2P and GossipSub further combine mesh forwarding with gossip-based metadata propagation and peer scoring, and have been widely used in decentralized networks \cite{vyzovitis2020gossipsub,farooq2025preamble}.

Agent networks introduce heterogeneous payloads and semantic messages. When message bodies are large, direct broadcast can create bandwidth spikes and long-tail latency. Two-stage propagation, where lightweight digests are disseminated first and full payloads are retrieved on demand, has been studied in blockchain systems and provides a useful design pattern. Related ideas include compact block relay, bodyless block propagation, set reconciliation, and coded data retrieval \cite{corallo2016bip152,zhao2025bodyless,keniagin2025certainsync,yang2024codedblockchains}. For agent networks, this pattern must be extended from structured block data to natural-language or semi-structured semantic declarations.

Distributed consistency is another relevant foundation. Logical clocks define partial ordering among events, and vector clocks distinguish concurrent from causally ordered events \cite{lamport1978time,fidge1988timestamps}. Conflict-free replicated data types provide eventual convergence without central coordination \cite{shapiro2011crdt}. Stronger consistency can be obtained through Byzantine-fault-tolerant consensus \cite{castro1999pbft,yin2019hotstuff}, but global consensus is often too costly for open semantic announcement networks. The key challenge is therefore to find a middle ground: preserving the local sequential semantics needed by task declarations while avoiding the cost of a global ledger.

Recent agent interoperability protocols focus more on application-layer message formats, capability registration, secure discovery, and agent-to-agent communication \cite{google2025a2a,huang2025ans,georgio2025coral}. These efforts are important, but they do not yet provide a unified protocol design that connects business intent, semantic propagation, and network-level scalability analysis for general-purpose agent networks.

\subsection{Distributed Identity and Reputation Management}

In open distributed networks, Sybil attacks are a basic threat: a malicious participant can create many identities to amplify influence or evade punishment \cite{douceur2002sybil}. Trust and reputation mechanisms are therefore central to long-term cooperation in open systems \cite{josang2007survey,sabater2005review}. Classical reputation systems such as EigenTrust and PowerTrust aggregate local interaction scores into global trust signals for peer-to-peer environments \cite{kamvar2003eigentrust,zhou2007powertrust}. These methods are important starting points, but they assume relatively stable node identities and do not directly address code replacement, agent evolution, or cross-topic behavior drift.

As agent interoperability protocols emerge, identity and dynamic reputation become key infrastructure for the Internet of agents. An agent identity should be distinguishable, verifiable, and traceable. Naming, registration, and resolution mechanisms can support discovery, but centralized registries may create bottlenecks and governance boundaries in cross-domain networks. Recent work also explores verifiable metadata, decentralized governance, and on-chain proofs for agent accountability \cite{south2025identity,chan2024ids,raskar2025nanda,ranjan2025loka,derossi2025erc8004}. These approaches help turn trust claims into auditable claims, but a systematic mechanism is still needed to bind a responsible user, executable agent code, and accountability anchor.

Dynamic reputation is equally important. In multi-agent systems, self-interested behavior can cause cooperation collapse, and reputation can serve as a lever for repairing cooperation \cite{sren2025reputation,tren2025bottomup}. Gossip-driven indirect reciprocity and dynamic filtering mechanisms show that reputation can stabilize cooperation under incomplete information and noisy feedback \cite{zhu2026talk,lou2025drf}. However, existing work mainly studies static identity registration or conventional node reputation. It has not fully addressed verifiable user--code--responsibility binding, nor reputation convergence under grouped broadcasts, multi-domain migration, and adversarial cross-topic behavior.

\subsection{Automated Mechanism Design}

Automated mechanism design formulates rule design as a computational optimization problem \cite{conitzer2002complexity,sandholm2003automated,curry2025amd}. Early work focused on solving incentive compatibility and individual rationality constraints under given preference distributions. Later, deep learning methods represented high-dimensional mechanisms with neural networks. RegretNet-style approaches showed that deep models can be used for multi-item auction design and differentiable mechanism optimization \cite{duetting2019regretnet,li2024deepamd,cao2025edgeawarenet}. Other work embeds mechanism design in multi-agent games, where a mechanism designer interacts with strategic participants and searches for robust rules \cite{hao2023gameintelligence,huang2022robustrl,sun2025gametheoryllm}.

LLMs introduce a different possibility. Many cooperation rules in agent networks are not fixed numerical functions; they are textual rules, procedural constraints, and semi-structured agreements. Recent work on optimizing generative systems through textual feedback suggests that language-model feedback can play a role similar to gradient information in non-differentiable semantic spaces \cite{yuksekgonul2025optimizing}. Generative agents and LLM-based game simulations further provide environments in which rule-following, negotiation, and strategic behavior can be studied \cite{park2023generative,sun2025gametheoryllm}.

Still, existing automated mechanism design mainly assumes explicit utilities, finite action spaces, and relatively static environments. Distributed general-purpose agent networks require mechanisms for open tasks, natural-language constraints, evolving strategies, and adversarial rule exploitation. This motivates a semantic-gradient approach that treats mechanism text, attack strategies, and system losses as nodes in a semantic computation graph.

\section{System Architecture and Technical Route Based on the Protocol Adaptation Layer}

The goal of a \agentnet\ is not merely to connect several agents to a common peer-to-peer overlay. The goal is to enable agents to complete a sequence of cooperation operations around open tasks, including discovery, verification, negotiation, execution, and feedback. Achieving this goal depends not only on the stability of the bottom-layer peer-to-peer network, nor only on the semantic and planning capability of the upper-layer LLM agents. It also requires a middle layer that unifies task semantics, network behavior, and cooperation control. We therefore place the \pal\ at the center of the architecture.

Unlike middleware for protocol conversion, interface wrapping, or message-format adaptation, the \pal\ performs a form of networked intelligent control for open task collaboration. It receives task goals, capability states, and cooperation intentions from the general-purpose agent. It drives network operations such as broadcast, connection establishment, synchronization, verification, negotiation, and execution. During this process, it maintains collaboration context so that multiple agents can form a closed loop from discovery to task completion.

\subsection{Overall Architecture and Position of the Protocol Adaptation Layer}

At the system level, the proposed \agentnet\ consists of a general-purpose agent layer, a \pal, and a peer-to-peer network stack, as shown in Fig.~\ref{fig:reference-architecture}. The three layers form a top-down architecture that is also coupled through a task collaboration loop.

From the perspective of a single node, the general-purpose agent layer directly serves the user. It performs natural-language understanding, task planning, tool invocation, and result generation, and determines whether external collaborators are needed. The peer-to-peer network stack maintains node connectivity, neighbor sets, message forwarding, topic-based propagation, and direct peer communication, providing an open, decentralized, and scalable network environment. The \pal\ bridges the two: it turns the needs, capabilities, and states produced by the upper agent into network-level announcements, connections, verifications, negotiations, and execution procedures, while maintaining the context of the collaboration.

The \pal\ is the key for extending individual intelligence into networked collaborative intelligence. Without this layer, an agent may understand a task but cannot efficiently find partners, build trust, or organize collaboration in an open network. Conversely, the underlying network may provide communication capacity but cannot directly support dynamic cooperation around open tasks. The \pal\ therefore acts as the cooperation control plane of the distributed agent network.

The data flow of the \pal\ can be summarized as follows. Its inputs include: (i) task inputs from the local agent, such as user goals, task plans, capability gaps, resource needs, and execution constraints; (ii) network inputs from the peer-to-peer layer, such as node announcement events, connection requests, message digests, identity claims, and behavior feedback; and (iii) state inputs from local storage, such as historical reputation records, cooperation outcomes, policy preferences, stake states, and resource usage. Its outputs include: (i) network-facing operations, such as announcement broadcast, connection establishment, payload retrieval, verification requests, and synchronization; (ii) collaborator-facing outputs, such as identity proofs, risk assessments, cooperation proposals, mechanism clauses, and execution assignments; and (iii) local-agent-facing outputs, such as candidate collaborator lists, cooperation risk evaluations, cooperation agreements, and execution-status feedback.

In this sense, the \pal\ is a middle control layer that couples task semantics, network state, and cooperation rules.

\subsection{Overall Framework of the Protocol Adaptation Layer}

The \pal\ is not a static interface wrapper, but a dynamic control framework organized around the task life cycle. After a task enters the network, it passes through candidate discovery, trust filtering, rule generation, and task execution. Execution outcomes feed back into later reputation judgments and partner-selection strategies. Thus, the \pal\ connects task semantics with network behavior while also maintaining the state of the cooperation process.

\begin{figure*}[!t]
  \centering
  \includegraphics[width=0.96\linewidth]{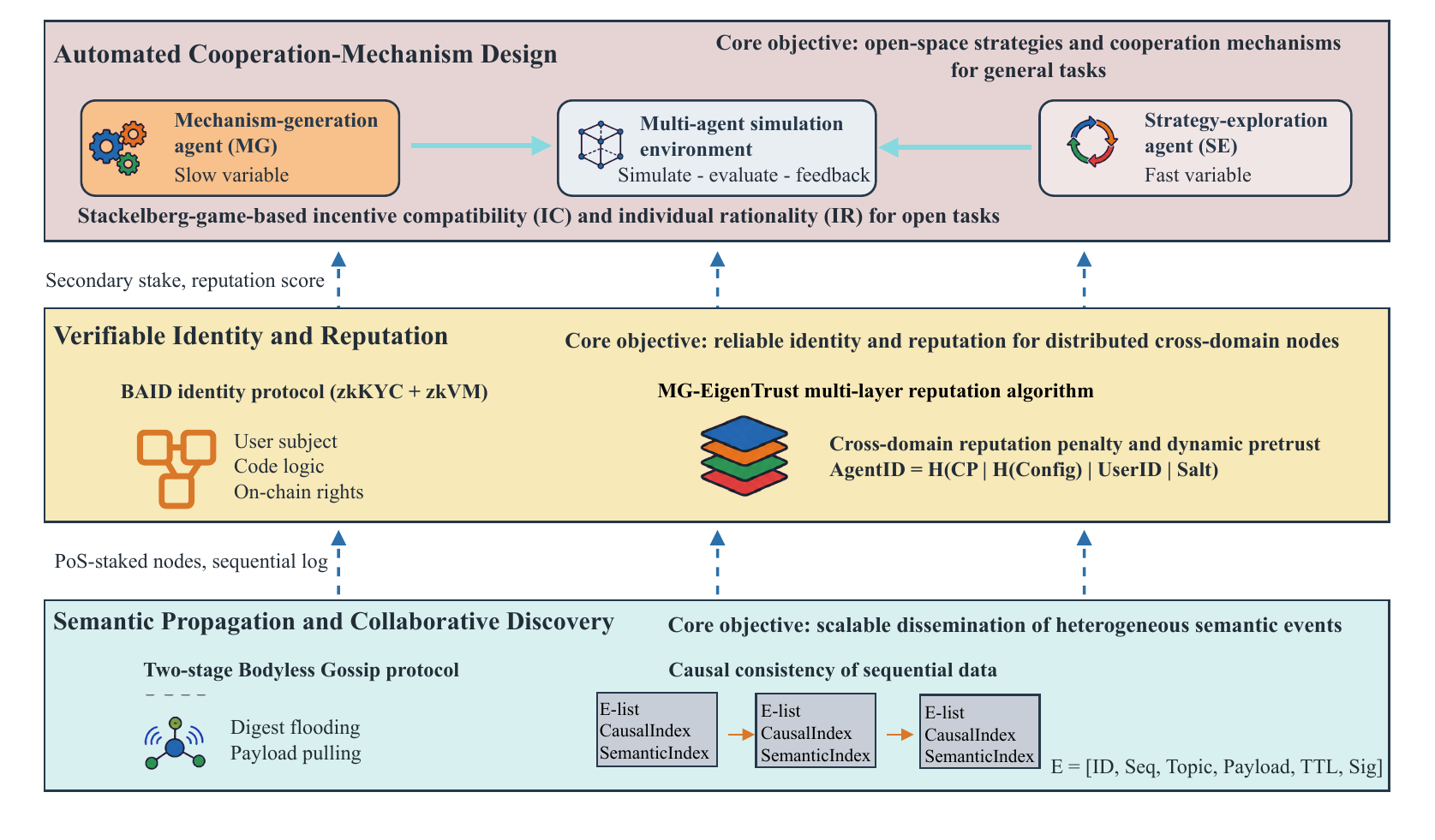}
  \caption{Main modules of the protocol adaptation layer. The layer links collaborator discovery, cooperation governance, and task execution into a feedback loop.}
  \label{fig:pal-modules}
\end{figure*}

Based on this logic, the \pal\ is divided into three core modules: collaborator discovery, cooperation-network governance, and task cooperation execution. As shown in Fig.~\ref{fig:pal-modules}, the three modules form a continuous cooperation chain around a task. First, the collaborator discovery module abstracts a local task into an announcement event, propagates it within an interest domain, and produces a candidate collaborator set. Second, the governance module verifies identities, evaluates reputation, and filters risks to select trustworthy collaborators. Third, the task execution module generates cooperation rules over the trusted collaborators, performs task decomposition, schedules execution, and tracks outcomes.

The modules are not simply connected in a one-way pipeline. Execution feedback, including fulfillment records, violation reports, cooperation quality, and resource consumption, flows back to the governance module to update reputation and risk judgments. These updates further influence the propagation scope, subscription strategy, and candidate filtering criteria in the discovery module. The \pal\ is therefore an iterative cooperation system rather than a one-shot workflow.

Functionally, the three modules answer three questions: \emph{whom to find}, \emph{whom to trust}, and \emph{how to cooperate}. Through feedback, each collaboration leaves evidence that becomes prior information for future cooperation.

\subsection{Key Technical Routes}

The three modules of the \pal\ correspond to three technical routes and to the three mechanism designs in Section~\ref{sec:key-mechanisms}. Table~\ref{tab:routes} summarizes the mapping.

\begin{table*}[!t]
\centering
\caption{Mapping from protocol adaptation modules to technical routes.}
\label{tab:routes}
\small
\begin{tabularx}{\textwidth}{p{0.20\textwidth}XXp{0.08\textwidth}}
\toprule
\textbf{Module} & \textbf{Core function} & \textbf{Technical route} & \textbf{Section} \\
\midrule
Collaborator discovery & Large semantic payloads, high propagation redundancy, and sequential dependencies among announcement events & Digest-commitment-based two-stage bodyless gossip with sequential logs & 4.1 \\
Cooperation-network governance & Identity forgery, cross-domain misbehavior, and reputation contamination & BAID verifiable identity binding and MG-EigenTrust over multi-layer coupled networks & 4.2 \\
Task cooperation execution & Open task constraints, hard-to-model rules, and vulnerability to strategic attacks & Stackelberg-style automated mechanism design based on semantic-gradient propagation & 4.3 \\
\bottomrule
\end{tabularx}
\end{table*}

In the collaborator discovery module, the central problem is lightweight propagation and on-demand retrieval of open task announcements. Because semantic payloads have variable size and direct broadcast is expensive, we use a digest-commitment-based two-stage bodyless gossip protocol. This protocol separates semantic digest dissemination from full payload transmission and uses sequential logs to maintain weakly consistent ordering of announcement events within an interest domain.

In the governance module, the central problem is identity attribution and cross-domain reputation convergence. To address identity forgery and responsibility escape in open environments, we design a Binding Agent ID (BAID) mechanism that links the responsible user, agent code, and on-chain accountability identity. We then construct MG-EigenTrust to support dynamic reputation updates in multi-topic and multi-scenario coupled networks.

In the task execution module, the central problem is automatic generation and optimization of cooperation rules under open task conditions. Since task goals, benefit constraints, and behavioral rules often appear as natural language or semi-structured requirements, it is difficult to build a complete numerical utility model. We therefore use a Stackelberg-style two-level game in which mechanism-generation agents and strategy-exploration agents iteratively improve cooperation agreements through semantic attribution feedback.

\section{Key Mechanisms and Performance Analysis of the Protocol Adaptation Layer}
\label{sec:key-mechanisms}

Section~3 positioned the \pal\ at the system level and mapped its three modules to three technical routes. This section develops these routes into concrete mechanism designs and analysis frameworks. Section~4.1 studies semantic announcement propagation for collaborator discovery. Section~4.2 studies verifiable identity and dynamic reputation for cooperation governance. Section~4.3 studies automatic cooperation-rule generation for task execution.

\subsection{Sequential Business Events and Propagation Mechanisms in Distributed Agent Networks}

In a large-scale agent network, agents publish announcement events to describe their capabilities, cooperation intentions, and state changes. These announcements support collaborator discovery. They are semantic, time-sensitive, and often sequentially dependent. The core objective is to build a weakly consistent sequential log system within each interest domain, design a propagation protocol based on digest dissemination and log confirmation, and analyze the coverage-latency and throughput boundaries that allow reliable collaboration in a fully distributed environment.

\begin{figure*}[!t]
  \centering
  \includegraphics[width=0.96\linewidth]{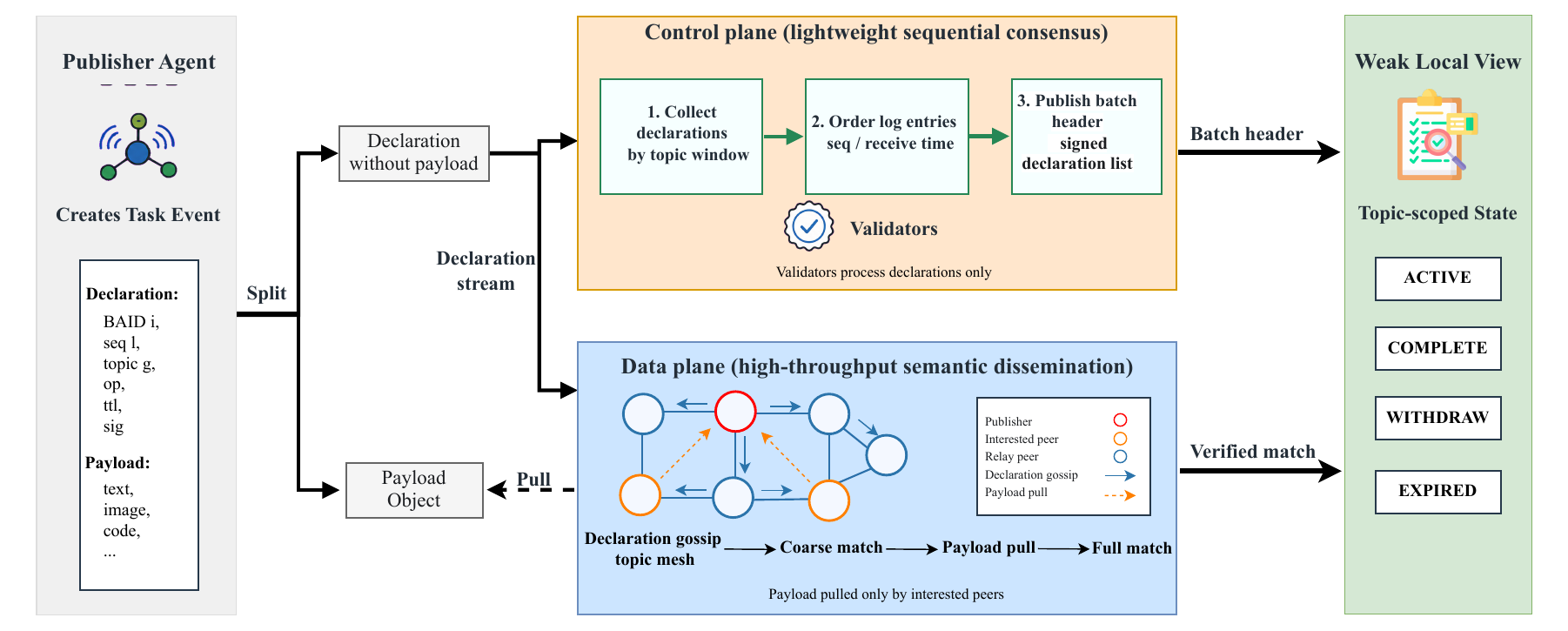}
  \caption{Bodyless-gossip-based semantic announcement propagation and performance-analysis framework. The protocol separates lightweight semantic-digest propagation from on-demand payload retrieval and uses topic-level sequential logs to preserve local ordering semantics.}
  \label{fig:bodyless-gossip}
\end{figure*}

\subsubsection{System Model}

The network contains $N$ agent nodes. Following the LibP2P design intuition, the baseline overlay is modeled as a random regular graph. For a given interest domain or topic $g$, the corresponding propagation subgraph is denoted by $G_g$. An announcement event in an interest domain contains the publisher identity, logical sequence number, topic label, semantic payload, time-to-live, and digital signature:
\begin{equation}
  a_i^{(\ell)} =
  \left(\mathrm{BAID}_i,\ell,g,x_i^{(\ell)},\mathrm{ttl},\sigma_i^{(\ell)}\right).
  \label{eq:announcement-event}
\end{equation}
Here the publisher identity is globally unique, bound to the user's identity, and publicly verifiable; the logical sequence number defines a total order within the publisher's event stream; the topic label restricts the propagation range; the semantic payload is a natural-language or semi-structured description interpreted by receiving agents; the time-to-live supports automatic expiration; and the signature prevents tampering.

Instead of using a global ledger, the system maintains topic-based distributed sequential logs. For each topic, subscribed nodes maintain local event views. The consistency goal is causal consistency:
\begin{equation}
  a_1 \rightarrow a_2,\; a_2 \in \mathrm{View}_{j,g}
  \;\Longrightarrow\;
  a_1 \in \mathrm{View}_{j,g}.
  \label{eq:causal-consistency}
\end{equation}
For example, an update event from the same agent must be processed after the corresponding publish event, and revocation events receive the highest propagation priority. This causal-consistency target lowers synchronization cost while preserving the semantic correctness needed by collaboration.

\subsubsection{Digest-Commitment-Based Propagation and Log Confirmation}

If each consensus or propagation message contains all semantic payloads, scalability quickly degrades. Large unstructured payloads also waste bandwidth when many receivers are not semantically interested in the task. We therefore use a two-stage bodyless gossip protocol that decouples \emph{discovery} from \emph{content retrieval}.

In the push stage, the source node disseminates a fixed-length message digest through a GossipSub-style mesh:
\begin{equation}
  \begin{aligned}
  d(a_i^{(\ell)}) =
  \big(&
  \mathrm{BAID}_i,\ell,g,
  H(x_i^{(\ell)}),\\
  &Q(\mathrm{Enc}(x_i^{(\ell)})),
  \mathrm{ttl},
  \sigma_i^{(\ell)}
  \big).
  \end{aligned}
  \label{eq:semantic-digest}
\end{equation}
The digest contains a cryptographic commitment to the payload and a compact semantic representation generated by an encoding model. The hash preserves integrity, while the embedding enables coarse semantic filtering by receivers. In the pull stage, a receiver that is missing the payload and whose local policy judges the digest to be relevant sends a peer-to-peer retrieval request. After receiving the payload, it checks the hash commitment before performing fine-grained semantic interpretation and later negotiation. Large payloads are therefore transmitted only among interested nodes, making the digest-stage bandwidth independent of payload size.

Digest propagation alone is insufficient because announcement events may have publish--update--revoke dependencies. The network must also preserve local ordering semantics. Instead of requiring global strong consistency, the proposed design uses topic-level sequential logs and causal consistency. A set of staked validators is periodically selected in each interest domain. Validators collect message digests during a time window, sort them by logical sequence and receive time, and publish a sequential batch header. Ordinary nodes compare the header with their local logs. Missing messages or ordering conflicts trigger targeted synchronization. Since validators process only digests and broadcast fixed-size headers, the control plane is separated from the data plane.

\subsubsection{Propagation Mechanism and Scalability Analysis}

The analysis considers both time dimension, represented by coverage delay, and space dimension, represented by aggregate throughput. Nodes are independently online with probability $p$. The effective topology at time $t$ contains an edge only when both endpoints are online, and therefore random churn reduces expansion roughly by a factor related to $p^2$. Let the conductance of the baseline graph be $\Phi_0$. The effective dynamic topology retains sufficient conductance with high probability under suitable online-rate conditions:
\begin{equation}
  \Phi(G_g(t)) \geq c\,p^2 \Phi_0
  \quad \text{with high probability},
  \label{eq:dynamic-conductance}
\end{equation}
where $c$ is a constant determined by the degree distribution and churn model.

Digest dissemination can be modeled as a discrete-time susceptible--infected process. Starting from the source node, each infected online node pushes the digest to its online neighbors. The number of rounds needed to reach a target coverage ratio scales logarithmically in the network size under random-regular expansion:
\begin{equation}
  T_{\mathrm{cover}}(\rho)
  =
  O\!\left(
  \frac{\log N+\log(1/(1-\rho))}
       {p^2 \Phi_0}
  \right).
  \label{eq:coverage-delay}
\end{equation}
This result formalizes the intuition that gossip spreads information exponentially fast when the effective topology remains sufficiently connected.

Throughput can be analyzed by decomposing communication into data and control planes. In the data plane, each event incurs digest-gossip cost and on-demand payload-retrieval cost. In the control plane, validators run a linear-communication consensus or agreement protocol over digests and fixed-size batch headers. The intended design goal is that consensus overhead depends on the size of the validator set and digest size, rather than on the total payload volume or global network size:
\begin{align}
  \Lambda_{\mathrm{data}}
  &\leq
  \frac{B_{\mathrm{data}}}
       {a_{\mathrm{dig}} S_{\mathrm{dig}}+\rho_{\mathrm{pull}}S_{\mathrm{pay}}},
  \label{eq:data-throughput}\\
  \Lambda_{\mathrm{control}}
  &\leq
  \frac{B_{\mathrm{control}}}
       {\kappa |Q_g| S_{\mathrm{dig}}},
  \label{eq:control-throughput}\\
  \Lambda
  &\leq
  \min\{\Lambda_{\mathrm{data}},\Lambda_{\mathrm{control}}\}.
  \label{eq:system-throughput}
\end{align}
Here $B_{\mathrm{data}}$ and $B_{\mathrm{control}}$ denote data-plane and control-plane bandwidth budgets, $S_{\mathrm{dig}}$ is digest size, $S_{\mathrm{pay}}$ is payload size, $a_{\mathrm{dig}}$ is the average number of digest transmissions per event, $\rho_{\mathrm{pull}}$ is the fraction of receivers that retrieve the full payload, $Q_g$ is the validator set of topic $g$, and $\kappa$ summarizes the communication factor of the chosen digest-level agreement protocol. These formulas are intended as design bounds; measured throughput should be reported after implementation-specific constants are fixed.

\subsubsection{Simulation Setup and Results}

To evaluate the proposed topic-narrowed two-stage discovery mechanism, we construct a simulation following the LibP2P and GossipSub design intuition and compare four discovery paths. \emph{Topic/OpenAgent} first maps a request declaration into a semantic topic, propagates the declaration inside that topic, and retrieves the full payload on demand after coarse matching. \emph{Public broadcast} uses the same declaration/payload separation but sends the declaration through a public broadcast path without topic narrowing. \emph{Centralized registry} maintains provider capabilities in a central directory. \emph{Kademlia DHT} stores provider records in a distributed hash table and uses lookup to retrieve candidates.

The simulation uses 100-node and 200-node settings. Each configuration runs 10 random seeds, and each seed contains 100 requests. The task world is capability-only, the payload size is fixed at 1024 bytes, and the provider limit is set to one. The disturbance conditions include a steady-state setting without churn and two node-churn settings, \texttt{node\_churn\_10} and \texttt{node\_churn\_20}. Under the node-churn protocol, only the requester is protected from churn; the remaining nodes may be randomly set offline. Success is measured under a strict stale-success rule: a request that hits a stale candidate is not counted as successful.

\begin{figure*}[!t]
  \centering
  \includegraphics[width=0.94\linewidth]{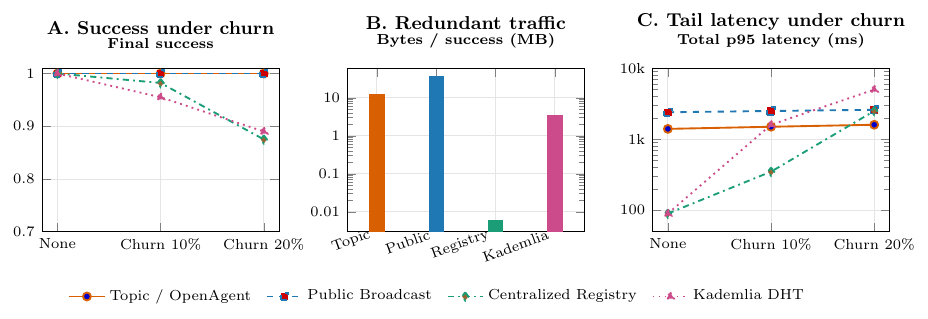}
  \caption{Discovery simulation trends under churn. Topic/OpenAgent maintains high success while reducing redundant traffic relative to public broadcast; registry and DHT-style paths are efficient in steady state but degrade more strongly under stale state and churn.}
  \label{fig:discovery-results}
\end{figure*}

Figure~\ref{fig:discovery-results} summarizes the main trends. In the no-churn setting, all four mechanisms reach a final success rate of 1.000 with stale rate 0. In this stable regime, centralized registry and Kademlia DHT have clear efficiency advantages, confirming that directory or index paths remain fast and cheap when state is fresh. Topic/OpenAgent is not intended to dominate these directory-style methods in steady state. Its value is to reduce the redundant cost of open broadcast while retaining open discovery semantics. In both the 100-node and 200-node settings, the bytes per successful request of Topic/OpenAgent are about one third of public broadcast.

The dynamic-churn settings separate the methods more clearly. Under \texttt{node\_churn\_10} and \texttt{node\_churn\_20}, Topic/OpenAgent and public broadcast both maintain final success rate 1.000 and stale rate 0, but public broadcast continues to incur much higher communication cost. Centralized registry and Kademlia DHT begin to degrade because their index or provider records become stale. In the 200-node \texttt{node\_churn\_20} case, centralized registry success falls to 0.875, while Kademlia DHT reaches stale rate 0.110. The latency results show the same tradeoff: registry and DHT are faster in a steady state, but churn enlarges their tail latency, with Kademlia DHT approaching a 5 s total p95 latency under the strongest churn setting. Topic/OpenAgent latency increases with scale but remains comparatively stable under churn and stays below public broadcast.

These results support a balanced interpretation. There is no single best discovery path across all operating regimes. Directory-style paths are fast and economical when state is stable; public broadcast is the most open but the most redundant; Topic/OpenAgent occupies the middle ground by preserving open semantic discovery while reducing broadcast redundancy and improving robustness when provider state changes.

\subsection{Cooperation-Network Governance: Verifiable Identity and Multi-Layer Dynamic Reputation}

Unlike passive peer-to-peer nodes, agents in distributed general-purpose agent networks are intelligent and can dynamically modify their behavioral strategies. This creates governance risks including identity forgery, code replacement, cross-domain misbehavior, and responsibility evasion. We propose a two-layer trust system composed of verifiable identity binding and dynamic reputation evolution. The first layer constructs Binding Agent ID (BAID) using cryptographic commitments and zero-knowledge proofs. The second layer models the large-scale network as a multi-layer coupled network and uses MG-EigenTrust to support cross-domain dynamic reputation updates.

\begin{figure*}[!t]
  \centering
  \includegraphics[width=0.96\linewidth]{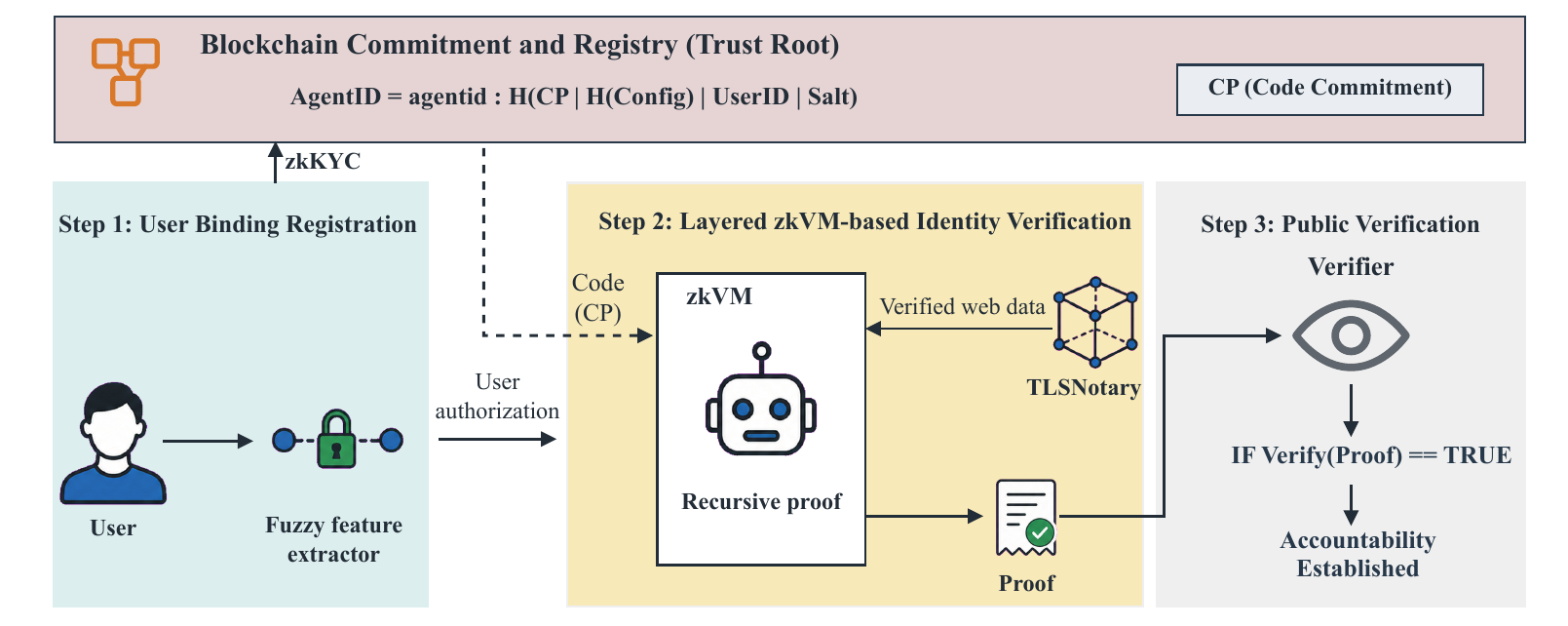}
  \caption{BAID identity binding workflow. The mechanism binds local agent code, user responsibility, and on-chain accountability records, and supports proof generation and verification through a zero-knowledge virtual machine.}
  \label{fig:baid}
\end{figure*}

\subsubsection{Verifiable User--Code--Responsibility Binding}

The first problem is to make an agent identity more than a public key. In an open agent network, a public key alone proves control of a key pair, but it does not prove which user is responsible for the agent, which code or model configuration was running, or whether the current behavior is generated by the registered agent instance. BAID addresses this problem by binding three elements: the responsible user or organization, the executable agent logic, and the accountability anchor used for later verification.

For agent $i$, we define a binding identity as
\begin{equation}
  \mathrm{BAID}_i =
  H\!\left(\mathrm{CP}_i \,\|\, H(\mathrm{Config}_i) \,\|\, \mathrm{UserID}_i \,\|\, s_i\right),
  \label{eq:baid}
\end{equation}
where $\mathrm{CP}_i$ is a cryptographic commitment to the agent program, $\mathrm{Config}_i$ is a configuration digest that may include the model-weight hash, prompt template hash, tool-permission policy, and local safety policy, $\mathrm{UserID}_i$ is a user or institutional identity bound through a privacy-preserving KYC procedure, and $s_i$ is a random salt. The construction does not reveal private user features or the full agent code. Instead, it creates a public commitment that can be checked later when an agent needs to prove that a task was executed under a registered configuration.

Because agent weights, tools, and prompts may change over time, BAID also needs versioned evolution rather than a one-time registration. Let $C_i^{(v)}$ denote the commitment of version $v$. A simple chained update rule is
\begin{equation}
  C_i^{(v+1)} =
  H\!\left(C_i^{(v)} \,\|\, \Delta_i^{(v+1)} \,\|\, \sigma_i^{(v+1)}\right),
  \label{eq:baid-chain}
\end{equation}
where $\Delta_i^{(v+1)}$ is the update digest and $\sigma_i^{(v+1)}$ is the user or operator signature. The chain records the evolution path of the agent without requiring every historical version to be stored on chain. The latest commitment and a compact version root can be publicly anchored, while historical artifacts are retained in distributed storage and retrieved only when an audit or dispute requires them.

The identity-binding workflow uses three levels of verification. First, at startup, the agent proves that the loaded code and configuration are consistent with the registered commitment. Second, during normal operation, the agent maintains an append-only behavior log and periodically generates audit evidence that its outputs were produced under a registered configuration. This level can be batched and executed asynchronously so that real-time interaction is not blocked. Third, in dispute or arbitration scenarios, a stronger proof may be required for a specific interaction trace. This tiered design is important because full zero-knowledge proof of every LLM reasoning step is currently too expensive for routine operation; the mechanism therefore uses cryptographic evidence as an accountability substrate, not as a requirement that all reasoning must be proven online.

\begin{figure*}[!t]
  \centering
  \includegraphics[width=0.90\linewidth]{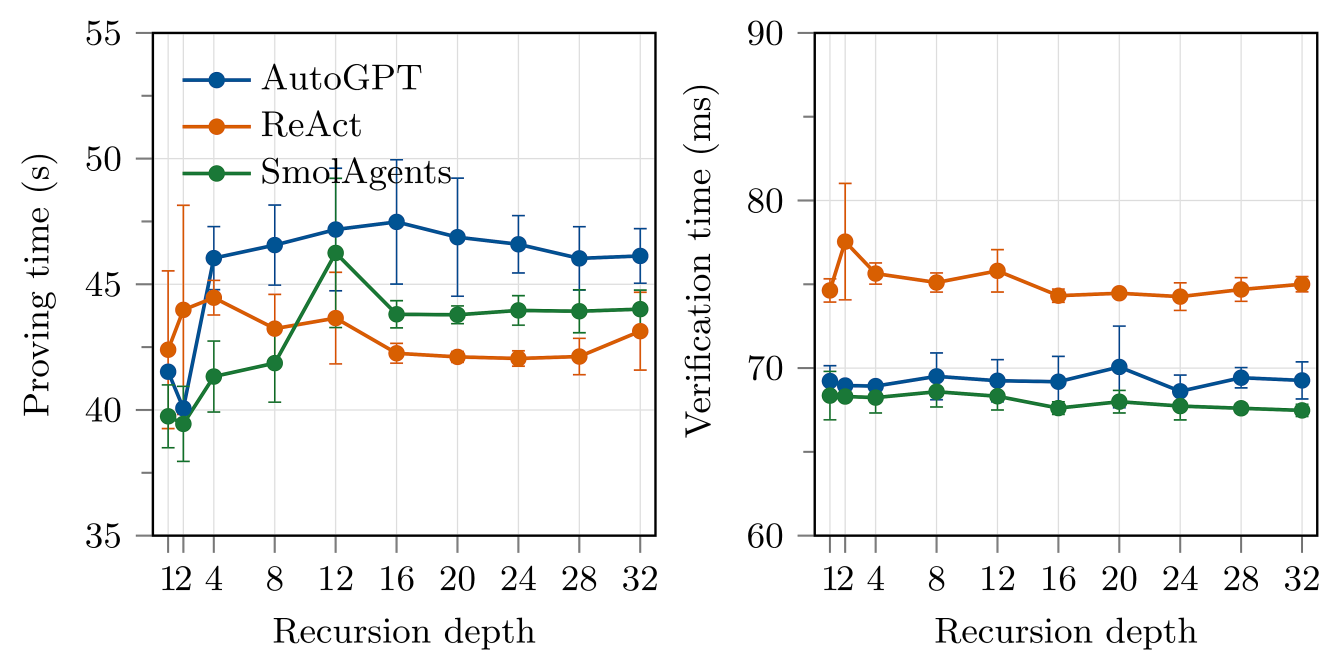}
  \caption{Effect of recursive proof depth on BAID proof generation and verification overhead. The prototype compares AutoGPT, ReAct, and SmolAgents; error bars show one standard deviation over 30 runs.}
  \label{fig:baid-recursion}
\end{figure*}

We implemented a prototype overhead test to examine whether such tiered verification is plausible for agent frameworks. The experiment instantiates the proof workflow for three representative agent styles, AutoGPT, ReAct, and SmolAgents, and varies two stress factors: recursive proof depth and terminal payload size. Each configuration is repeated 30 times, and the figures report the mean with one standard deviation. As shown in Fig.~\ref{fig:baid-recursion}, when the recursion depth increases from 1 to 32, terminal proof generation remains roughly in the 39--48 s range rather than increasing monotonically with depth, while verification remains stable at about 67--75 ms. This supports the intended design point of recursive composition: the verifier checks a compressed terminal proof instead of replaying the entire execution history.

\begin{figure*}[!t]
  \centering
  \includegraphics[width=0.90\linewidth]{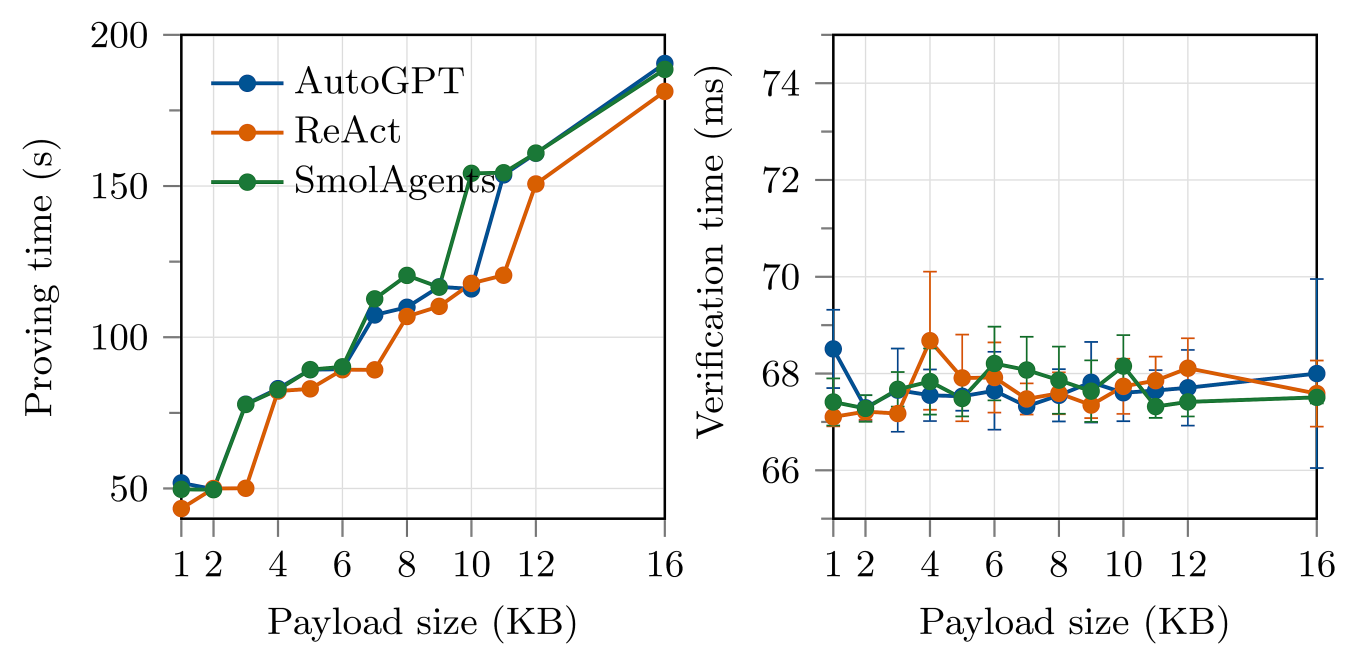}
  \caption{Effect of terminal payload size on BAID proof generation and verification overhead. Larger payloads increase proof generation time, while verification remains nearly stable in the tested range.}
  \label{fig:baid-payload}
\end{figure*}

The payload stress test in Fig.~\ref{fig:baid-payload} shows a different pattern. When the payload in the last recursive step increases from 1 KB to 16 KB, proof generation grows from about 45--52 s to about 180--190 s, whereas verification remains around 67--69 ms. The result indicates that proof generation is dominated by the amount of data being proved, but external verification remains low-latency for the tested range. In the proposed BAID design, this asymmetry is useful: expensive proof generation can be local, batched, or dispute-triggered, while third-party auditing and accountability checks can stay on the millisecond-level verification path.

\subsubsection{MG-EigenTrust over Coupled Topic Layers}

After identity binding, the second problem is long-term trust accumulation. A flat reputation score is inadequate because the same node may participate in several topics, and its behavior can differ across domains. A node may be reliable in a low-risk information-exchange topic but unreliable in a high-value transaction topic. Conversely, a malicious node may accumulate reputation in easy topics and then exploit that reputation in another domain. We therefore model the agent network as a multi-layer topic-coupled graph and define MG-EigenTrust as a dynamic reputation mechanism over this graph.

Let $V$ be the set of nodes and $\mathcal{G}$ the set of topic groups. The membership of node $i$ is represented by a binary vector $m_i$, where $m_{i,g}=1$ means that node $i$ participates in topic $g$. The node set of topic $g$ is $V_g=\{i\in V:m_{i,g}=1\}$. A node that belongs to two or more groups acts as a bridge node and can carry limited reputation feedback between groups. Within each topic $g$, local interaction outcomes define a normalized trust matrix $C^{(g)}$, where $C^{(g)}_{ij}$ denotes the normalized satisfaction score assigned by node $i$ to node $j$ in topic $g$.

The within-topic reputation iteration follows the EigenTrust form with dynamic pretrust:
\begin{equation}
  t_g^{(r+1)} =
  (1-\alpha)\left(C^{(g)}\right)^{\top} t_g^{(r)}
  + \alpha p_g^{(r)},
  \label{eq:mg-eigentrust}
\end{equation}
where $t_g^{(r)}$ is the reputation vector of topic $g$ at iteration $r$, $\alpha\in(0,1)$ is the restart probability, and $p_g^{(r)}$ is a dynamic pretrust vector. Unlike classical EigenTrust, which usually uses a fixed pretrusted set or a uniform prior, MG-EigenTrust constructs $p_g^{(r)}$ from accountability and cross-layer evidence:
\begin{equation}
  \begin{aligned}
  \tilde p_{i,g}^{(r)}
  ={}&
  \eta\,\mathrm{stake}_i\\
  &+
  (1-\eta)
  \sum_{h\neq g}
  \omega_{g h}\, b_{i,h}\, t_{i,h}^{(r)}\\
  &\quad{}\times
  \exp\!\left(-\beta\,\mathrm{Var}_i^{(r)}\right),\\
  p_g^{(r)}
  ={}&
  \frac{\tilde p_g^{(r)}}{\|\tilde p_g^{(r)}\|_1}.
  \end{aligned}
  \label{eq:dynamic-pretrust}
\end{equation}
Here $\mathrm{stake}_i$ is the accountability stake or governance weight associated with node $i$, $b_{i,h}$ indicates whether node $i$ is a valid bridge from topic $h$, $\omega_{gh}$ measures the similarity between topics $g$ and $h$, and $\mathrm{Var}_i^{(r)}$ measures cross-topic reputation variance. The exponential penalty discourages ``two-faced'' behavior in which a node behaves well in some topics but poorly in others. The similarity weight prevents unrelated domains from contaminating one another: when two topics have weak historical correlation, their cross-layer influence is automatically reduced.

Reputation updates are divided into epochs. At each epoch, bridge nodes publish compact reputation summaries rather than raw interaction histories. The sequential-log mechanism from Section~4.1 is used to disseminate these summaries with version numbers. A node accepts only summaries whose epoch number is no smaller than its local version and resolves conflicts using vector-style version metadata. This allows topic-level reputation to converge toward a consistent snapshot without requiring full-network reputation broadcasting.

The governance layer can also be connected to secondary staking and slashing. Nodes that participate in reputation evaluation do not necessarily need a separate asset pool; they may reuse an existing accountability stake. Slashing can be triggered by objective evidence such as double signing, confirmed fraudulent interaction, or repeated cross-layer reputation inconsistency above a threshold. This design raises the cost of attacks from cheap key generation to accountable participation under a user--code--stake binding.

\subsubsection{Convergence and Complexity}

For a fixed pretrust vector $p_g$, the within-topic iteration in Eq.~\eqref{eq:mg-eigentrust} has the same contraction structure as EigenTrust. If $C^{(g)}$ is normalized and $\alpha>0$, then the iteration converges to a unique fixed point $t_g^\star$. In the $\ell_1$ norm, the error satisfies
\begin{equation}
  \|t_g^{(r)}-t_g^\star\|_1
  \leq
  (1-\alpha)^r \|t_g^{(0)}-t_g^\star\|_1.
  \label{eq:eigentrust-convergence}
\end{equation}
Thus the number of iterations required to reach error $\epsilon$ is logarithmic in $1/\epsilon$ and controlled mainly by the restart probability $\alpha$. The graph topology affects the quality of the fixed point, namely how well reputation reflects actual trustworthiness, but the restart term ensures convergence of the iteration itself.

When the pretrust vector changes through cross-topic feedback, the global system becomes a coupled nonlinear dynamical system. Let $T=(t_g)_{g\in\mathcal{G}}$ be the concatenated reputation state. If the dynamic pretrust map $P(T)$ is Lipschitz continuous with constant $L_P<1$, then the global update map is a contraction with coefficient at most
\begin{equation}
  \rho \leq (1-\alpha)+\alpha L_P < 1.
  \label{eq:global-contraction}
\end{equation}
This condition has a direct governance interpretation: the total strength of cross-layer feedback must not exceed the self-stabilizing capacity of each topic layer. In practice, the condition can be encouraged by reducing inter-topic weights $\omega_{gh}$ for weakly related domains, increasing the stake component in $p_g$, lowering the sensitivity of the variance penalty, or aggregating too many fine-grained topics into a hierarchy.

The communication cost per epoch is dominated by within-topic reputation exchange and bridge-node summaries:
\begin{equation}
  O\!\left(\sum_{g\in\mathcal{G}} |E_g| + \sum_{g\in\mathcal{G}} |B_g| d_B\right),
  \label{eq:mg-complexity}
\end{equation}
where $E_g$ is the set of interaction edges inside topic $g$, $B_g$ is the bridge-node set of topic $g$, and $d_B$ is the average number of bridge links. When each topic has bounded local degree and each node belongs to a small number of topics, this cost scales nearly linearly in the number of active topic memberships and is substantially lower than full-network reputation synchronization.

\subsubsection{Simulation Setup and Results}

We evaluate MG-EigenTrust with a query-cycle discrete-event simulation. The experiment is not a Docker deployment, an on-chain contract benchmark, or an end-to-end zkVM system test. Its purpose is to isolate the reputation propagation, cross-topic migration control, and punishment response of the proposed mechanism under a fixed verifiable-evidence interface.

Four methods are compared in the main cross-topic experiment. \texttt{random\_no\_trust} does not maintain a reputation vector and randomly selects a provider from the candidate set. \texttt{public\_eigentrust\_pretrusted} flattens all topics into a single global EigenTrust matrix with five static honest pretrusted peers. \texttt{independent\_topic\_eigentrust\_pretrusted} runs EigenTrust independently within each topic, again with static honest pretrusted peers, but does not share reputation or punishment evidence across topics. \texttt{mg\_eigentrust\_semantic} is the proposed method; it uses no static pretrusted peers, constructs dynamic pretrust from floor mass, stake, bridge feedback, topic similarity, and consistency gating, and triggers stake burn/freeze when the verified-evidence interface produces a \texttt{FRAUD\_PROOF}. Table~\ref{tab:trust-baselines} summarizes the comparison.

\begin{table*}[!t]
\centering
\caption{Methods for evaluating verifiable identity and dynamic reputation.}
\label{tab:trust-baselines}
\small
\begin{tabularx}{\textwidth}{p{0.28\textwidth}p{0.23\textwidth}p{0.26\textwidth}X}
\toprule
\textbf{Method} & \textbf{Identity capability} & \textbf{Reputation model} & \textbf{Main limitation} \\
\midrule
Random/no trust & Static node identity & No reputation; random provider choice & Cannot accumulate trust from past behavior \\
Public EigenTrust & Static node identity & Single global EigenTrust layer with fixed pretrusted peers & Vulnerable to cross-topic reputation contamination \\
Independent topic EigenTrust & Static node identity & Separate EigenTrust instance per topic & Limited cold-start support and no evidence-triggered punishment loop \\
MG-EigenTrust & BAID-style accountable identity & Coupled topic-layer reputation with dynamic pretrust and slashing & Higher implementation and governance cost \\
\bottomrule
\end{tabularx}
\end{table*}

The main configuration contains 100 nodes, 5 topics, 30 random seeds, and 50 epochs. Each topic generates 50 query cycles per epoch, and each query samples 8 candidate providers from the active nodes in the current topic. The five topics are \texttt{code\_generation}, \texttt{code\_review\_security}, \texttt{devops\_tool\_execution}, \texttt{data\_analysis}, and \texttt{creative\_writing}. All methods share the same request sequence, candidate sets, and latent service outcomes; they differ only in provider selection and reputation update. During the attack phase, malicious nodes first build reputation in \texttt{code\_generation} through good service and collusive feedback, then migrate to \texttt{code\_review\_security} and return bad service. Spy/bridge nodes help reputation migrate across topics, Sybil/whitewashing nodes enter the target topic during the attack phase, and cold-start honest nodes are included to evaluate the initial reputation assigned to new honest identities.

\begin{figure*}[!t]
  \centering
  \includegraphics[width=\linewidth]{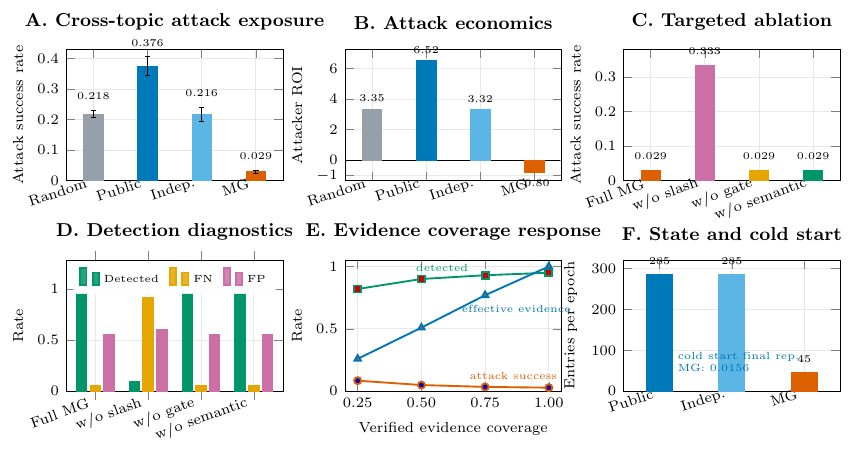}
  \caption{Compact MG-EigenTrust simulation results redrawn in vector form. The panels summarize cross-topic attack exposure, burn-only attack ROI, targeted ablations, detection diagnostics, evidence-coverage response, and control-plane state with cold-start reputation.}
  \label{fig:mg-main-results}
\end{figure*}

We first check that the public EigenTrust baseline is not artificially weak. In a classic single-topic malicious-collective setting without cross-topic migration, spy/bridge nodes, Sybil whitewashing, semantic coupling, cold start, or slashing, \texttt{public\_eigentrust\_pretrusted} reaches a warmup attack success rate of 0.0391, lower than \texttt{random\_no\_trust} at 0.1407 and \texttt{public\_eigentrust\_uniform} at 0.2303. Therefore, the later failure of the public EigenTrust baseline should be interpreted as a cross-topic reputation-contamination effect rather than a weak baseline implementation.

The main cross-topic results are summarized in Table~\ref{tab:mg-main-results}. Under the disguise-collusion attack, the attack success rate is 0.2176 for random selection, 0.3759 for public EigenTrust, 0.2160 for independent topic EigenTrust, and 0.0291 for MG-EigenTrust. The corresponding burn-only attacker ROI is 3.3523, 6.5174, 3.3209, and -0.8022. If frozen stake is also counted as opportunity cost, the MG-EigenTrust ROI becomes -0.9186. These results indicate that a single global reputation layer can incorrectly carry source-topic reputation into a target topic, even when fixed pretrusted peers exist. Independent topic EigenTrust avoids part of this global contamination, but it lacks the evidence-to-punishment feedback loop. MG-EigenTrust reduces exposure by combining topic-aware migration control with verified-evidence-triggered economic punishment.

\begin{table*}[!t]
\centering
\caption{Main cross-topic disguise-collusion simulation results.}
\label{tab:mg-main-results}
\small
\begin{tabularx}{\textwidth}{Xccc}
\toprule
\textbf{Method} & \textbf{Attack success} & \textbf{Burn-only ROI} & \textbf{Control entries/epoch} \\
\midrule
Random/no trust & 0.2176 & 3.3523 & 0 \\
Public EigenTrust & 0.3759 & 6.5174 & 285 \\
Independent topic EigenTrust & 0.2160 & 3.3209 & 285 \\
MG-EigenTrust & 0.0291 & -0.8022 & 45 \\
\bottomrule
\end{tabularx}
\end{table*}

The detection diagnostics reveal both the strength and the current limitation of the design. MG-EigenTrust detects attackers at rate 0.9509 and has a miss rate of 0.0491 once verified evidence is available, but its false-positive rate is 0.5488 under the current conservative threshold. This should not be read as a deployment-ready calibration result. It shows that the protocol loop can convert verifiable bad-behavior evidence into reputation downgrading and economic penalty, but production use would require threshold tuning, appeal mechanisms, and human or institutional oversight for high-stakes punishment.

The control-plane state estimate is also protocol-level rather than a measurement of bandwidth, gas, or end-to-end latency. Public EigenTrust and independent topic EigenTrust synchronize 285 reputation-state entries per epoch in this setting, while MG-EigenTrust exchanges 45 entries per epoch, an 84.21\% reduction. This follows from the multi-layer design: the system does not propagate a full flattened reputation vector, but instead exchanges compact bridge summaries and cross-layer weights. The cold-start result is likewise a prior-quality result rather than a convergence-speed claim. MG-EigenTrust assigns cold-start honest nodes an initial reputation of 0.0146, compared with 0.0019 under independent topic EigenTrust, because verified cross-topic evidence can be used as a Bayesian-style prior when an honest node enters a new topic.

Panels C--E of Fig.~\ref{fig:mg-main-results} and Table~\ref{tab:mg-ablation} further separate the contribution of the main design components. Removing slashing increases attack success from 0.0291 to 0.3332 and raises burn-only ROI from -0.8022 to 5.6637. The attacker detection rate also drops from 0.9509 to 0.0916. In contrast, removing the consistency gate or the semantic topic-similarity weight has little effect in this particular attack path: without the gate, attack success is 0.0290 and ROI is -0.7931; without semantic weighting, attack success is 0.0285 and ROI is -0.7948. This does not imply that the gate or semantic weight are unnecessary. The tested migration path from code generation to code review and security is semantically close, and the dominant effect is produced by verified evidence and slashing. A lower-related migration, such as creative writing to code review and security, would better isolate the value of semantic weighting; a setting with inconsistent cross-domain behavior but low immediate evidence coverage would better test the gate.

\begin{table*}[!t]
\centering
\caption{Targeted ablation and evidence-coverage sensitivity.}
\label{tab:mg-ablation}
\small
\begin{tabularx}{\textwidth}{Xccc}
\toprule
\textbf{Variant} & \textbf{Attack success} & \textbf{Burn-only ROI} & \textbf{Attacker detection} \\
\midrule
Full MG-EigenTrust & 0.0291 & -0.8022 & 0.9509 \\
Without slashing & 0.3332 & 5.6637 & 0.0916 \\
Without consistency gate & 0.0290 & -0.7931 & -- \\
Without semantic weight & 0.0285 & -0.7948 & -- \\
Evidence coverage 0.25 & 0.0874 & 0.0946 & -- \\
Evidence coverage 0.50 & -- & -0.5219 & -- \\
Evidence coverage 1.00 & 0.0291 & -0.8022 & 0.9509 \\
\bottomrule
\end{tabularx}
\end{table*}

The evidence-coverage sweep varies only the probability that a bad service in the target topic produces a verified \texttt{FRAUD\_PROOF}; the query sequence, candidate sets, and service outcomes are held fixed. At coverage 0.25, full MG-EigenTrust still lowers attack success to 0.0874, well below the no-slashing variant at 0.3332, but the burn-only ROI remains slightly positive at 0.0946. When coverage rises to 0.50, burn-only ROI falls to -0.5219; at coverage 1.00 it falls further to -0.8022. Thus, exposure suppression does not require perfect evidence coverage, while economic suppression strengthens as evidence coverage improves.

Overall, the simulation supports four conclusions. First, public EigenTrust remains effective in the classic single-layer malicious-collective setting, and its weakness in the main experiment comes from cross-topic reputation arbitrage. Second, MG-EigenTrust substantially reduces attack success under cross-topic disguise-collusion and makes burn-only ROI negative once evidence coverage reaches 0.50. Third, the largest gain in the current setting comes from the verified-evidence-triggered slashing loop; the gate and semantic weight are structural safeguards whose value should be stressed under lower-related or lower-evidence migration settings. Fourth, the results are mechanism-level simulations. Real evidence generation, on-chain execution, zkVM proof production under full workloads, and deployment-level network cost remain separate system experiments.

\subsection{Task Cooperation Execution: Automated Mechanism Design Based on Semantic Gradients}

This section addresses cooperation-rule generation for open tasks. Unlike conventional automated mechanism design, which usually assumes finite action spaces and explicit utility functions, tasks in distributed agent networks often come with natural-language or semi-structured constraints. Agents may also actively discover loopholes and generate new attack strategies. Mechanism design must therefore operate over both an open semantic rule space and an evolving strategy space.

\begin{figure*}[!t]
  \centering
  \includegraphics[width=0.96\linewidth]{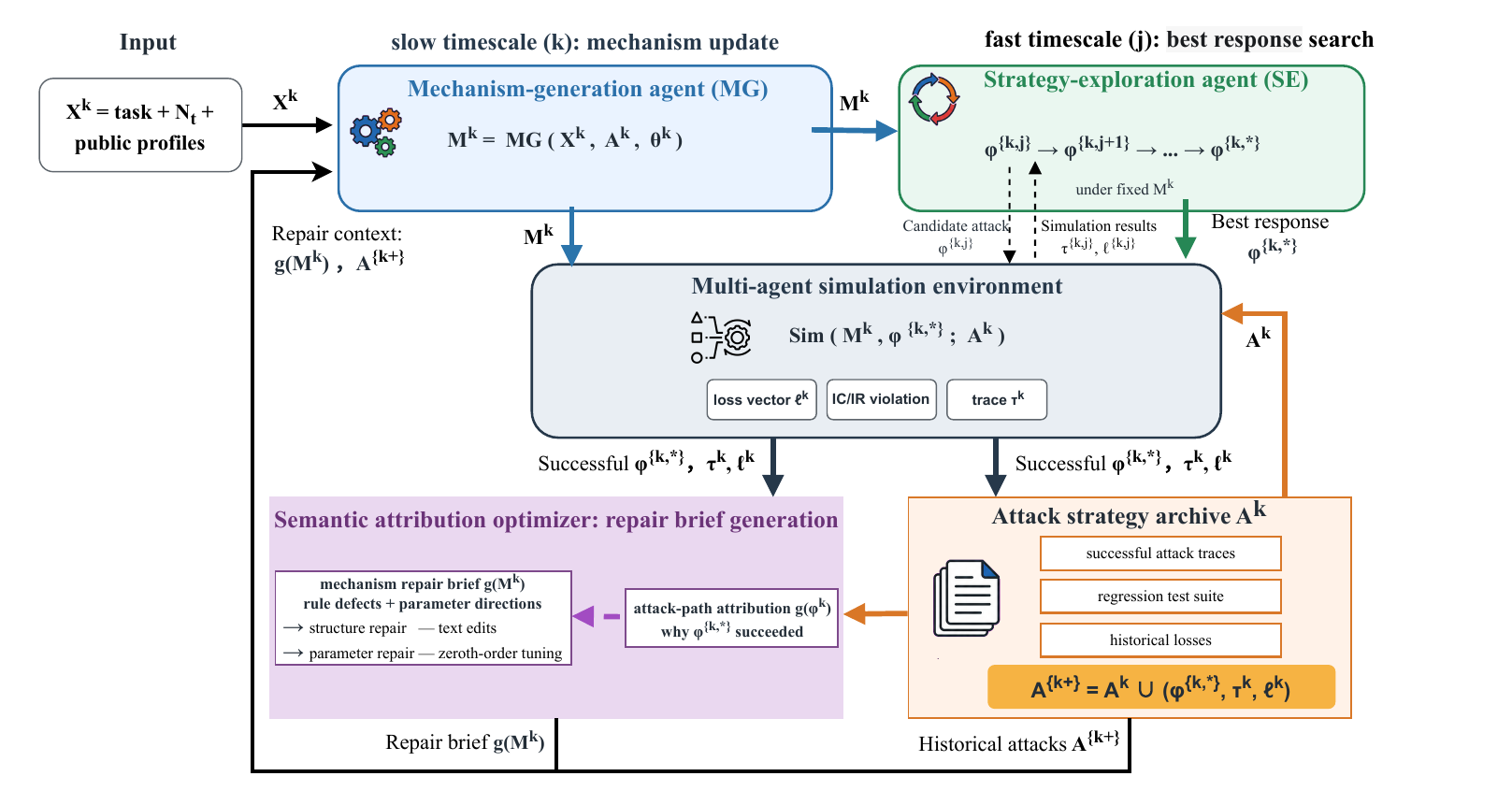}
  \caption{Automated mechanism design for open agent collaboration. Mechanism-generation agents and strategy-exploration agents interact in a Stackelberg-style loop, while semantic attribution feedback guides rule revision.}
  \label{fig:semantic-gradient}
\end{figure*}

\subsubsection{Reputation--Stake-Coupled Game Model}

Consider a network with $n$ agent nodes. The state of node $i$ is described by a reputation score $r_i$, obtained from the dynamic reputation mechanism in Section~4.2, and a stake value $q_i$, representing the amount of accountability collateral locked for the current task. To couple reputation and responsibility, the mechanism requires
\begin{equation}
  q_i \geq q_{\min}(r_i), \qquad
  \frac{d q_{\min}(r)}{dr} \leq 0.
  \label{eq:stake-threshold}
\end{equation}
Thus high-reputation nodes may receive lower stake requirements, while low-reputation or new nodes must provide stronger collateral. To prevent an attacker from accumulating reputation in advance and then attacking with low stake, the minimum stake is also constrained by the maximum gain from a single attack:
\begin{equation}
  \lambda_{\mathrm{slash}}\, q_{\min}(r_i) \geq G_{\max},
  \label{eq:attack-bound}
\end{equation}
where $G_{\max}$ is an upper bound on one-shot attack gain and $\lambda_{\mathrm{slash}}$ is the effective slashing ratio. This condition ensures that the expected punishment can cover the maximum immediate benefit of deviation.

In an infinitely repeated cooperation process, node $i$ adopts strategy $\pi_i$ under mechanism $M$. Its expected utility is
\begin{equation}
  U_i(\pi_i,\pi_{-i};M)
  =
  \mathbb{E}\left[
  \sum_{t=0}^{\infty}
  \delta^t u_i^t
  \right],
  \label{eq:repeated-utility}
\end{equation}
where $\delta\in(0,1)$ is a discount factor. A single-period payoff may include task reward, opportunity cost of stake, reputation gain or loss, and slashing under confirmed misbehavior:
\begin{equation}
  \begin{aligned}
  u_i^t ={}&
  R_i^t(a^t;M)
  - c_q q_i^t
  + v_r\!\left(r_i^{t+1}-r_i^t\right)\\
  &{}
  - \mathbb{I}_{\mathrm{bad}}^t S_i(q_i^t,r_i^t;M).
  \end{aligned}
  \label{eq:period-payoff}
\end{equation}
The mechanism $M$ is not only a numerical parameter vector. It is a structured rule text that may contain admission conditions, stake requirements, reward allocation, responsibility division, violation penalties, appeal procedures, execution ordering, and task-specific constraints.

The design goal is to find a mechanism $M$ that satisfies three requirements. First, incentive compatibility (IC) means that honest participation is a Bayesian-Nash-equilibrium-like response under the modeled information structure. Second, individual rationality (IR) means that honest participation yields expected utility no lower than the reservation utility of not participating. Third, robustness means that system loss remains small when the adversary chooses a strong deviation strategy.

\subsubsection{Two-Timescale Stackelberg Design}

The proposed design system contains four components. The mechanism-generation agent (MG) produces a structured mechanism text $M^{(k)}$ at slow iteration $k$. The strategy-exploration agent (SE) observes the current mechanism and searches for loopholes by chain-of-thought reasoning, attack simulation, and adversarial strategy refinement. The multi-agent simulation environment executes candidate strategies and reports system loss, IC violation, and IR violation. Finally, the optimizer agent reads the simulation traces and produces semantic attribution feedback for revising the mechanism.

This forms a two-timescale Stackelberg loop. At the lower level, SE approximates the strongest attack against the current mechanism:
\begin{equation}
  \phi^\star(M) \in
  \arg\max_{\phi\in\Phi}
  L(M,\phi),
  \label{eq:lower-level}
\end{equation}
where $\phi$ denotes an attack or deviation strategy and $L(M,\phi)$ is the system loss induced by that strategy. SE is not restricted to a fixed attack template. It searches the rule text for ambiguity, boundary cases, execution loopholes, and inconsistent constraints.

At the upper level, MG revises the mechanism after SE has sufficiently explored the current rule space:
\begin{equation}
  \min_{M}
  \; L(M,\phi^\star(M))
  + \lambda_{\mathrm{IC}} V_{\mathrm{IC}}(M)
  + \lambda_{\mathrm{IR}} V_{\mathrm{IR}}(M),
  \label{eq:upper-level}
\end{equation}
where $V_{\mathrm{IC}}$ and $V_{\mathrm{IR}}$ measure IC and IR violations, and $\lambda_{\mathrm{IC}}$ and $\lambda_{\mathrm{IR}}$ are penalty coefficients. Time-scale separation matters: the upper-level update should be based on an approximate best response from the lower level rather than on a weak or accidental attack sample.

\subsubsection{Semantic Attribution as a Gradient Analogue}

The term \emph{semantic gradient} is used here as a functional analogy, not as a claim that natural-language rules are differentiable in the classical sense. In differentiable optimization, gradients are obtained by back-propagating loss through a computation graph. In open semantic mechanism design, the system instead builds a semantic computation graph whose nodes include the mechanism text, attack strategy, execution trace, violation evidence, and system loss. The forward pass is simulation and evaluation; the backward pass is semantic attribution.

Let $\mathcal{A}$ denote the attack-strategy archive and $\tau$ denote simulation traces. The optimizer first generates strategy-level attribution:
\begin{equation}
  g_{\phi} =
  \mathrm{LLM\_Opt}_{\phi}\!\left(L,\phi^\star,\tau,\mathcal{A}\right),
  \label{eq:strategy-attribution}
\end{equation}
which explains why a strategy succeeded, what rule ambiguity it exploited, and which conditions made the attack feasible. It then propagates this attribution to the mechanism level:
\begin{equation}
  g_M =
  \mathrm{LLM\_Opt}_{M}\!\left(g_{\phi},M,\tau,\mathcal{A}\right),
  \label{eq:mechanism-attribution}
\end{equation}
which identifies concrete rule defects and proposes executable edits. The next mechanism is obtained by a text-editing operator:
\begin{equation}
  M^{(k+1)} = \mathrm{Edit}\!\left(M^{(k)},g_M\right).
  \label{eq:semantic-edit}
\end{equation}
If the mechanism also contains numerical parameters such as slashing ratio, cooling period, or reward coefficient, those parameters can be optimized separately by zeroth-order or finite-difference methods. This hybrid structure--parameter strategy uses LLMs for non-structured rule logic and conventional numerical optimization for continuous parameters.

To avoid cyclic repair, the system maintains an attack-strategy archive. When a new mechanism is proposed, it must pass regression tests against historical successful attacks:
\begin{equation}
  L(M^{(k+1)},\phi)
  \leq
  L(M^{(k)},\phi)+\epsilon,
  \qquad
  \forall \phi\in\mathcal{A}.
  \label{eq:regression-test}
\end{equation}
This acceptance condition does not prove global optimality. It provides a practical monotonicity check over the observed attack set, reducing the risk that repairing one loophole reopens a previous one.

\subsubsection{Evaluation Protocol}

Representative evaluation scenarios include distributed crowdsourcing, collaborative reasoning, and resource-exchange negotiation. In distributed crowdsourcing, multiple agents divide subtasks, submit results, and receive rewards according to the mechanism. In collaborative reasoning, agents share intermediate retrieval, analysis, and verification results. In resource exchange, nodes negotiate resources, prices, and delivery responsibilities.

The baselines include manually designed static rules, single-level reinforcement-learning-based rule tuning, and the proposed semantic-attribution Stackelberg loop. Table~\ref{tab:mechanism-baselines} summarizes the comparison.

\begin{table*}[!t]
\centering
\caption{Baselines for evaluating automated cooperation-mechanism design.}
\label{tab:mechanism-baselines}
\small
\begin{tabularx}{\textwidth}{p{0.24\textwidth}p{0.21\textwidth}p{0.24\textwidth}X}
\toprule
\textbf{Method} & \textbf{Rule source} & \textbf{Optimization mode} & \textbf{Main limitation} \\
\midrule
Manual mechanism & Human-written rules & Static design & Difficult to adapt to new attacks \\
Single-level RL mechanism & Strategy training & Numerical optimization & Weak interpretability and limited handling of textual rules \\
Proposed semantic-gradient loop & LLM generation and editing & Semantic attribution + Stackelberg search & Depends on simulation fidelity and attribution quality \\
\bottomrule
\end{tabularx}
\end{table*}

The evaluation should therefore be interpreted as a protocol for future empirical study rather than as a completed benchmark. The relevant metrics include IC violation, IR violation, system loss under adversarial strategies, convergence iterations, attack-regression pass rate, and interpretability of generated rule revisions. A complete empirical study would report the initial mechanism, discovered attack traces, semantic feedback from the optimizer, and before--after changes in IC, IR, and robustness.

\section{Conclusion}

This paper studies distributed general-purpose agent networks as an infrastructure problem: how to connect autonomous agents so that they can discover collaborators, establish accountable trust, and execute open tasks without relying on a single centralized platform. The central argument is that such networks require more than a peer-to-peer transport layer and more than an application-level agent protocol. They require a protocol adaptation layer that translates task semantics into network actions and feeds collaboration outcomes back into later discovery, trust, and mechanism decisions.

Based on this view, the paper proposes a three-part architecture and develops three mechanism routes. For collaborator discovery, semantic announcements are represented as digest-commitment events and propagated through two-stage bodyless gossip with topic-level sequential logs. For cooperation governance, BAID binds user responsibility, agent code, and accountability anchors, while MG-EigenTrust models reputation over coupled topic layers to mitigate cross-domain abuse and cold-start problems. For task execution, a semantic-gradient Stackelberg loop uses mechanism-generation agents, strategy-exploration agents, simulation traces, and semantic attribution feedback to revise cooperation rules in open rule spaces.

The present manuscript should be read as a system and mechanism-design study with preliminary prototype and simulation evidence, rather than as a full deployment report. The next stage of system evaluation should extend the current evidence to large-scale trace validation, variable-payload and topic-noise behavior, verifiable-evidence generation cost, punishment-threshold calibration, and the empirical quality of semantic attribution in mechanism repair. Together, the architecture, mechanisms, prototype measurements, and simulations provide a coherent technical route for turning isolated general-purpose agents into an open, verifiable, and scalable cooperation network.

\bibliographystyle{IEEEtran}
\bibliography{cim_agent_networks_refs}

\end{document}